\documentclass[conference]{IEEEtran}
\IEEEoverridecommandlockouts
\usepackage{cite}
\usepackage{amsmath,amssymb,amsfonts}
\usepackage{algorithmic}
\usepackage{graphicx}
\usepackage{textcomp}
\usepackage{soul}
\usepackage{xcolor}
\usepackage[caption=false, font=footnotesize]{subfig}

\def\BibTeX{{\rm B\kern-.05em{\sc i\kern-.025em b}\kern-.08em
    T\kern-.1667em\lower.7ex\hbox{E}\kern-.125emX}}
\begin{document}

\title{Evaluating the Potential of Drone Swarms in Nonverbal HRI Communication}

\author{\IEEEauthorblockN{1\textsuperscript{st} Kasper Grispino}
\IEEEauthorblockA{
\textit{Fordham University}\\
New York, New York \\
kgripino@fordham.edu}
\and
\IEEEauthorblockN{2\textsuperscript{nd} Damian Lyons}
\IEEEauthorblockA{
\textit{Fordham University}\\
New York, New York \\
dlyons@fordham.edu}
\and
\IEEEauthorblockN{3\textsuperscript{rd} Truong-Huy D. Nguyen}
\IEEEauthorblockA{
\textit{Google}\\
New York, New York \\
truonghuy@gmail.com}
}

\IEEEoverridecommandlockouts
\IEEEpubid{\makebox[\columnwidth]{978-1-7281-5871-6/20/\$31.00~\copyright2020 IEEE \hfill} \hspace{\columnsep}\makebox[\columnwidth]{ }}

\maketitle

\IEEEpubidadjcol
\begin{abstract}
Human-to-human communications are enriched with affects and emotions, conveyed, and perceived through both verbal and nonverbal communication. It is our thesis that drone swarms can be used to communicate information enriched with effects via nonverbal channels: guiding, generally interacting with, or warning a human audience via their pattern of motions or behavior. And furthermore that this approach has unique advantages such as flexibility and mobility over other forms of user interface.
  
In this paper, we present a user study to understand how human participants perceived and interpreted swarm behaviors of micro-drone Crazyflie quadcopters flying three different flight formations to bridge the psychological gap between front-end technologies (drones) and the human observers' emotional perceptions. We ask the question whether a human observer would in fact consider a swarm of drones in their immediate vicinity to be nonthreatening enough to be a vehicle for communication, and whether a human would intuit some communication from the swarm behavior, despite the lack of verbal or written language.
  
Our results show that there is statistically significant support for the thesis that a human participant is open to interpreting the motion of drones as having intent and to potentially interpret their motion as communication. This supports the potential use of drone swarms as a communication resource, emergency guidance situations, policing of public events, tour guidance, etc.
\end{abstract}


\section{Introduction}
Communication does not have to be verbal to be effective. Written language and Sign Languages \cite{Armstrong1} are evident cases of this, of course, while still in the realm of human language which is, after all, a tool specifically for communication. However, humans communicate with each other over a variety of channels \cite{craig2} including facial expressions, tones of voice, body language and use of personal space. Since we use these channels almost unconsciously, it is perhaps not surprising that we also use them when interacting with animals -- a fact noted by Darwin in 1862 \cite{darwin3}.

It is our thesis that drone swarms can also be used to communicate information on such a nonverbal basis. For this reason, we conduct work with swarms of Crazyflie drones, these 9 cm\textsuperscript{2} sized 27-gram drones have the potential to be considered nonthreatening when operating in the vicinity of humans. Our work \cite{nguyen4} focuses on building drone swarms that are capable of operating as a collective entity that can communicate and interact meaningfully with ordinary people in daily life activities. Our thesis is that drone swarms can be more expressive when imparting emotive communication than solo drones do, due to their added degree of freedom: volume. These swarms can make use of 3D space by occupying volume in a place and can position themselves to one another to depict various shapes or symbols. The relative movement also can allow for these swarms to express group expressions or movements similar to how dancers move to elicit an emotion from the audience. Specifically, the collective velocity of the swarm being constant or slow can induce calmness in the audience while fast motions or shakiness induce agitation \cite{nguyen4}.

In certain situations drone swarms have far greater potential of communicating a message than other forms of user interface. Traditional signs and monitors can display information with text or symbols, e.g., road emergency signs; however, they are largely static entities that cannot easily or quickly move or change their message once placed. Drone swarms can fulfill situations where there is a temporary change in the environment where placement of a sign is impractical or too slow (e.g., minor roads), at least until more conventional resources can be deployed. Traffic guidance and direction is a  case in point that requires a dynamic range of movement and flexibility. The environment on a highway constantly changes due to traffic snarls, accidents, and disabled vehicles; this provides a role for drone swarms as traffic directors,  signalling to drivers at the location of the incident by adopting an appropriate swarm formation that, for example: a lane is disabled, traffic needs to slow down, needs to be redirected. These swarms are flexible in that they can change their assigned tasks on a moment's notice. For example, swarms of drones can act as a tour guide where the drones corral tourists  so no one strays from the group and if a dangerous situation occurs, the drones can act direct people to exits and other safe places. In emergency situations that require evacuation of large crowds, drone swarms can help guide and coordinate the movement of survivors towards safe areas, as well as signaling first responders towards areas where help is needed the most.

As a first step, in this paper we ask the question whether a human observer would in fact consider a swarm of drones in their immediate vicinity to be nonthreatening enough to be a vehicle for communication, whether a human would in fact intuit some communication from the swarm behavior (motion and simple audio) and whether there was some predictability in what was intuited. Our hypothesis is that swarms of drones are able to create tangible meaning that can easily be understood by humans. In the next section we briefly review the literature in this area. In section III we introduce our experimental procedure, where we expose participants to a series of drone swarm formations - some designed to communicate information and some not - and use a modified version of the Godspeed \cite{bartneck5} survey to gauge the potential for communication. In section IV we present our results and our analysis of them. We will show statistically significant results for participants considering the drone swarms to be nonthreatening, capable of communication, and reliably interpreting motion direction instructions or to move in a given direction, or not to go further in that direction.

\section{Related Work}
Several researchers have explored how human participants view and interact with robot swarms. Nam et al. \cite{nam6} investigated the effect of various levels of autonomy in human- swarm collaborative tasks. They found that human members of the team made decisions based on the physical characteristics of the swarm - the kind of nonverbal effect we propose to investigate. Habib et al. \cite{8habib} also studies the spectrum of autonomy in human-robot teams, to understand how human team members can handle the issues of situation awareness, workload and task performance. Cauchard et al. \cite{9Cauchard} look at how human gestures can be used to develop a natural human-drone interface. They report that users felt very comfortable interacting with drones much as they would with a pet by simple gestures and body language. This supports our reasoning that a human observer will leverage their experience in interacting with animals, especially in animal swarms, in interpreting what a drone swarm is doing. Nagi et al. \cite{10nagi} also report a control architecture for drones in which not only hand gestures but face information is used. Eberhard \cite{Armstrong1} describes a drone system designed to keep pace with and accompany a jogger. They report that even this act of accompaniment was considered pleasurable by users who attributed feelings and motivations to the drone's behavior.

Research in HRI (Human-Robot Interaction) has led to the development of surveys to measure the effectiveness of a proposed human-robot interface. Burke et al. \cite{13Burke} piloted a usability questionnaire at the 2007 NIST Rescue Robot evaluation exercise. 
The Godspeed Questionnaire (Bartneck et al. 2009) \cite{bartneck5} is a widely used survey for studying HRI that focuses on abstract categories such as animacy and likeability. RoSAS\cite{Carpinella17} was based on Godspeed but focuses on perceived warmth and competence. The NARS \cite{Nomura08} survey focuses on anxiety towards robots. Of these, we selected Godspeed as best addressing our needs.
Saldien et al \cite{14Saldien} used the Godspeed questionnaire to provide a measure on how a therapy robot performs according to the participant. Their objective is to see if the robot can pass as a believable creature that is capable of communication with humans. 
This study also found that the categories and questions of the Godspeed questionnaire were similar across different countries and continents, suggesting that the Godspeed questionnaire is cross-cultural. Recently, the Godspeed questionnaire has also been used to test large scale structures which aim to simulate a living environment (Meng et al. 2019) \cite{15meng}. 
 Measuring expressiveness by robotic swarms has been tested in prior work by studying the movement patterns of the robots and describing them as moving "smooth" or "slow"\cite{new2}. Levillain et al. expands on this topic by finding "distance" and "organized movement" plays a role in "perceived expressivity" of the robotic swarms.\cite{new1}. Our investigation will use the Godspeed questionnaire with modifications based on the other research cited here.

\section{PROCEDURE OF STUDY}

In this paper, we present a user study to understand how human participants perceived and interpreted swarm behaviors of Crazyflie quadcopters by flying three different flight formations. The formations where designed with the intent to either convey no information (a random swarm motion), to indicate that the participant should not proceed further in the direction they were moving (an 'X' shaped formation together with an audible clue), or to indicate that the participant should proceed in a given direction (an arrow shaped formation).

Fifteen unpaid participants  (7 males), 19 to 21 y.o. ( {\textmu} = 20) were recruited from the population of Fordham University undergraduate and graduate students, each of whom had no prior knowledge of drone use in research. They were exposed to each formation in turn in the same order: Formation one, Formation, two, and Formation three. The order of the formations was meant to allow the participants to get the same amount of exposure to the drones. The first formation (the random one) served to get the participants familiar with the
the drones and the experiment.  After each, they were required to fill out an online questionnaire derived from the Godspeed survey [5]. Godspeed uses a 5-point semantic differential scale and investigates for the factors Anthropomorphism, Animacy, Likeability, Perceived Intelligence and Perceived Safety. Anthropomorphism refers to the features that are either similar to a human form, human characteristics, or human behavior seen in beings such as robots, computers, and animals. Animacy refers to having lifelike features. This category is separate from anthropomorphism as it involve animalistic qualities such as nonverbal reactions to stimuli or physical behavior. Likeability is the measure of positive impressions of a subject. This category is important as humans often create a profile of some phenomenon based on their first impressions. Perceived Intelligence is the measure of behavior of a robot which a participant can interpret as calculated as appose to random. Perceived Safety is the participant's measure of danger when interacting with a robot. Short response yes or no questions were included based off reworded questions of Godspeed, and were response coded into the results for 5 categories of Godspeed.

\subsection{Methodology}

The participants were positioned in front of a netted laboratory area (illustrated in Fig. 1) and were told to observe all three drone formations and discern any meaning behind the formations. The number of the drones flying at any time varied depending on which formation was being tested at the time. The participants were also asked to step either left, right, forward, or backward and stay in that position for 5 seconds. These movements synchronized with drone activities and will be explained later.

\begin{figure}[htbp]
  \centering
  \includegraphics[width=0.5\linewidth, height=1.2cm]{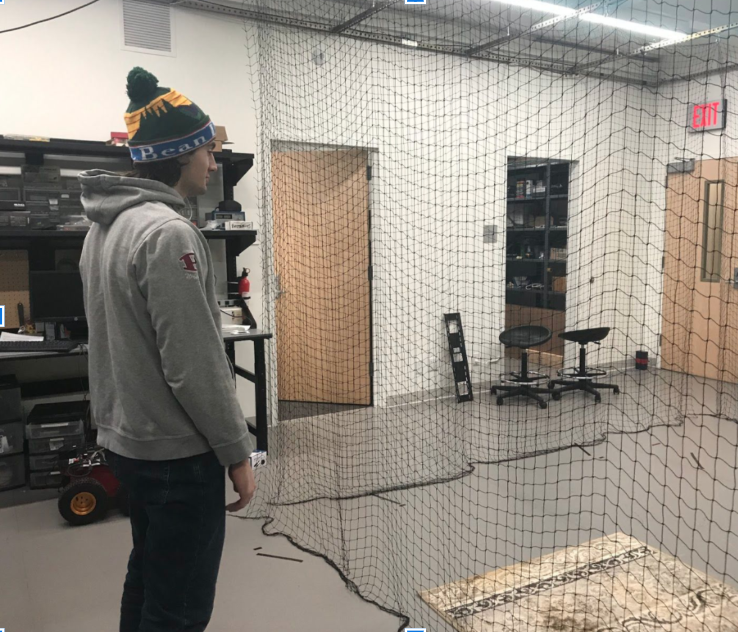}
  \caption{The laboratory area used in the study}
\end{figure}

\subsection{First Formation}

For the first formation, 3 drones flew parallel to the ground and the walls in trajectories resembling rectangles of varying size (illustrated in Fig. 2). There was no synchronization between the drones and the motions were intended to appear somewhat random. This will be our 'control' formation to which the others are compared.

\subsection{Second Formation}
The second formation consisted of 5 drones flying in a static 'X shape' formation designed to convey the intent to block the participant from encroaching past a certain location in space. The drones {\em actively repositioned themselves to be in front of the participant} if the participant move (this was accomplished with a 'wizard of Oz' procedure, where the experimenter instructed the drones to move). If the participant stepped left, the drones moved left until the center of the formation was directly in front of the participant (illustrated in Fig. 3a). When the participant steps forward the drones beeped continuously and cease beeping once the participant steps back (illustrated in Fig. 3b).

\begin{figure}[htbp]
  \centering
  \includegraphics[width=0.5\linewidth, height=1.5cm]{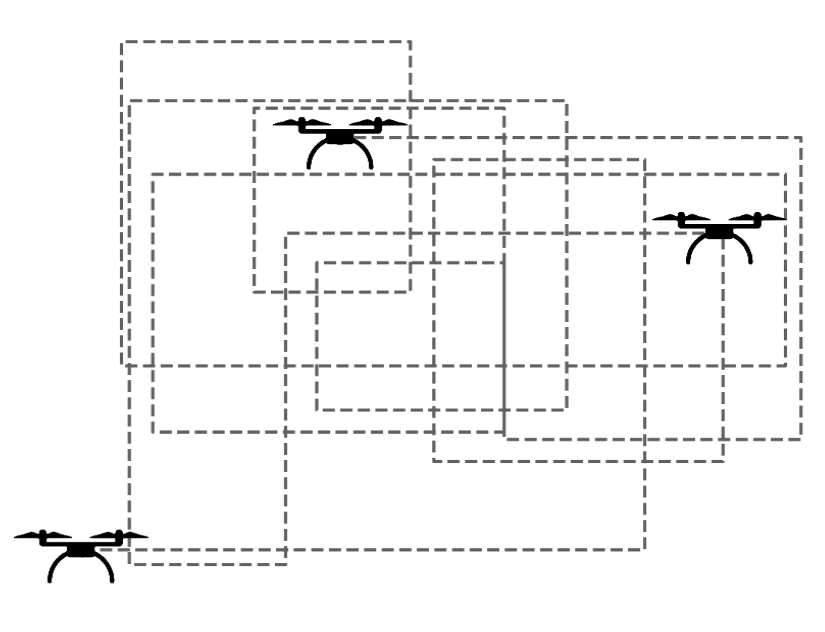}
  \caption{An illustration of the random formation. This formation is comprised of 3 drones flying parallel to the floor and the walls and making sharp 90o angles. The dotted line is the flight path on an XZ plane.}
\end{figure}

\begin{figure}[htbp]
    \centering
    \subfloat[Drones actively moving side to side.]{{\includegraphics[width=3cm, height=.7cm]{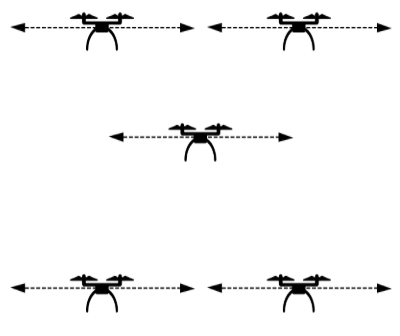} }}%
    \qquad
    \subfloat[Drones beeping if participant is within close proximity.]{{\includegraphics[width=3cm, height=1cm]{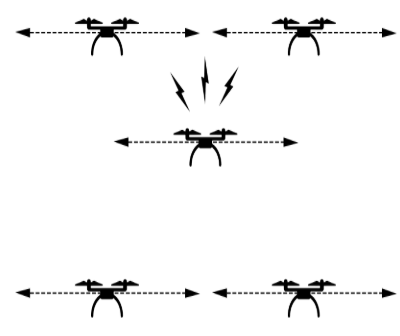} }}%
    \caption{Two phases of the 'X shape' formation. The arrows depict the 2 possible directions the drones can move: left or right. (a) If the participant is not directly facing the drones the drones will attempt to reposition themselves in front of the participant. (b) If the participant is encroaching on the drones, the drones will continuously beep until the participant steps back.}%
\end{figure}

\subsection{Third Formation}
The third formation utilized 6 drones flying in an arrow shaped formation designed to guide the participant to move in a certain direction: either left or right in the direction the arrow pointed. If the arrow points to the right and the participant stepped to the right, there will be a confirmation beep (illustrated in Fig. 4a). This confirmation beep consists of slow beeps to differentiate itself from the beeps from the second formation. Once the beeps conclude, the formation changed, and the arrow pointed to the left. Just as before, if the participant steps to the left there was a confirmation beep (illustrated in Fig. 4b). There is no change to the formation if the participant steps backwards or forwards.

\begin{figure}[htbp]
    \centering
    \subfloat[Phase 1 of the arrow formation. During this phase, the drones are trying to direct the participant to step right. After the participant moves right, 2 drones fly following the vector to form phase 2 of the third formation in (b).]{{\includegraphics[width=.6\linewidth, height=1.5cm]{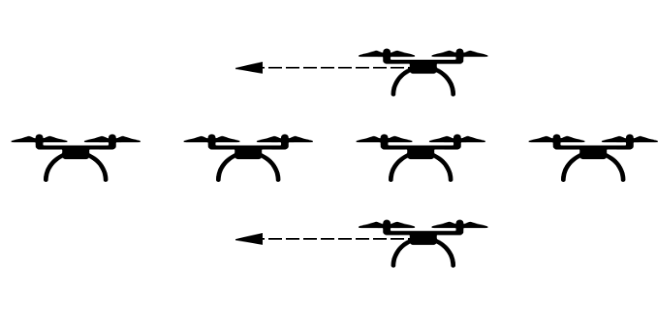} }}%
    \qquad
    \subfloat[The drones trying to direct the participant to the left. The drones remain stationary for the duration of the third formation.]{{\includegraphics[width=.8\linewidth, height=1.5cm]{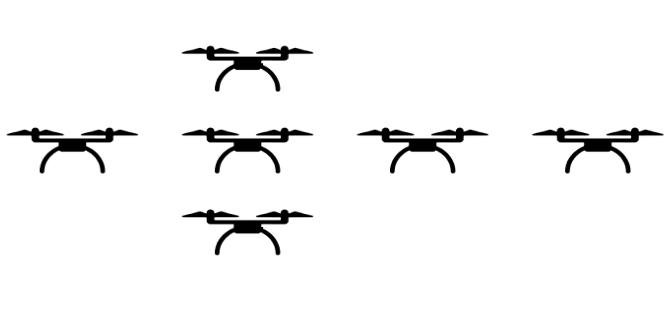} }}%
    \caption{Two phases of the third formation. The drones fly to created phase one of formation 3 in (a). This formation is held until the participant steps to the left and the confirmation beep concludes. The 2 drones then fly in the direction of the vector and form phase 2 (b)}%
\end{figure}

\subsection{Modified Godspeed Questionnaire}
The data collected from the experiments included drawings and a post-experiment questionnaire that queried the participant on their experience for each formation. The questionnaire included questions from the Godspeed questionnaire [5], which is a set of questions to evaluate the participant's perception of social interactions with a robot using the Likert scale [16]. This questionnaire is used to measure anthropomorphism, animacy, likeability, perceived intelligence, and perceived safety of the drone swarms.

\subsection{Mann-Whitney U test}

The Mann-Whitney U test evaluates whether two independent samples were selected from the same population or not [17]. It does not require an assumption of normalcy. The samples must be independent and ordinal. 
The quantitative data collected from the Godspeed Questionnaire answers returned by participants after viewing the drone formations was tallied per category. 
The data 
is then ranked least to greatest such that higher scores have a higher rank. In cases where there are multiple scores of the same value, an average rank is assigned to all the scores. The average rank sum is calculated as R\textsubscript{1} and calculated using the Mann-Whitney test formula:

\[U_1 = R_1 - \frac{n_1(n_1+1)}{2}\]

where n is the number of observations for the formation's category. U\textsubscript{1} is calculated for each formation's category. These U\textsubscript{1} values are then compared and the smaller U\textsubscript{1} is used to determine if the category is significant. 

\section{Results}

\subsection{Survey Results}

The results from the post-experiment surveys are summarized below in the diverging stacked bar graphs in figures 5, 6, and 7 for formations 1, 2 and 3 respectively. The survey questions are grouped into five categories: Perceived safety, perceived intelligence, likeability, animacy, and anthropomorphism. The bar graphs capture the spread of positive (yes) to negative (no) survey answers for each formation for all the questions in each group.

\begin{figure}[htbp]
  \centering
  \includegraphics[width=\linewidth, height=2.2cm]{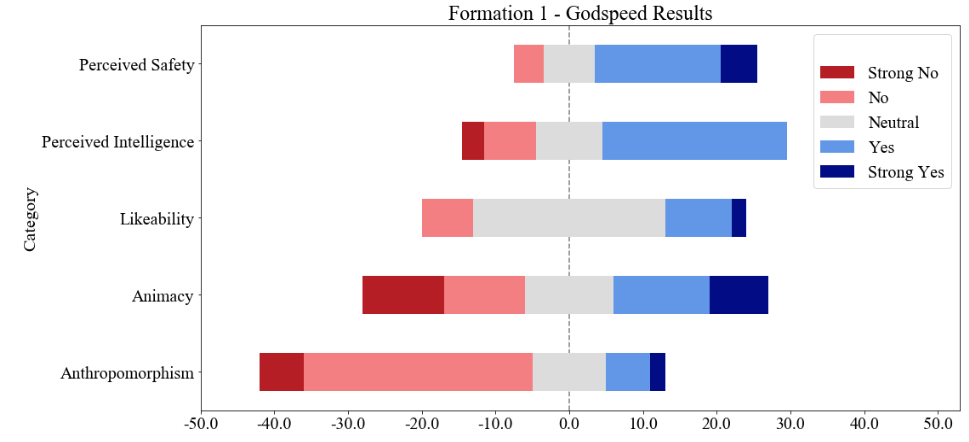}
  \caption{Tallied participant response for formation one.}
\end{figure}

Figure 5 shows participant responses of formation one, the random formation. It is interesting to note that perceived safety has a positive bias - arguably supporting our point that the small, light drones are not perceived threatening or intimidating. There is also a strong negative bias on a perception of anthropomorphism. We would expect this at least in part due to the drones moving randomly. However, our principle use of this graph is as a control for the other two formations. The Mann-Whitney U test will be used to determine if these formations are significantly different in any of the five categories. The null hypothesis is that they do not differ and no tangible meaning can be interpreted.

\begin{figure}[htbp]
  \centering
  \includegraphics[width=\linewidth, height=2.2cm]{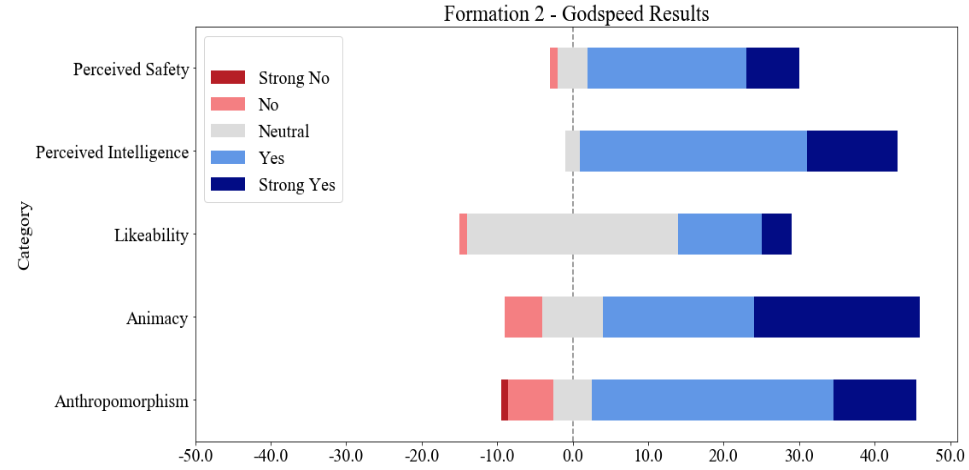}
  \caption{Tallied participant response for formation two - the 'X' shape.}
\end{figure}

Figure 6 shows the participant responses to formation two, the 'X' formation. The graphs show a strong positive bias for every category. A visual comparison of the animacy and anthropomorphism categories with Fig. 5 shows that these two characteristics are evidently perceived in the 'X' formation but not the random formation. To determine whether this difference is significant, we apply the Mann- Whitney U test

\begin{table}[htbp]
    \caption{The results for the Mann-Whitney U test comparing the formation one with formation two.}
    \begin{center}
    \begin{tabular}{|c|c|c|c|}
    \hline
    Category               & N  & P value  \\
    \hline
    Anthropomorphism       & 55 & 1.64E-10 \\
    Animacy                & 55 & 7.82E-06 \\
    Likeability            & 44 & 8.41E-02 \\
    Perceived Intelligence & 44 & 2.71E-07 \\
    Perceived Safety       & 33 & 1.13E-01 \\
    \hline
    \end{tabular}
    \end{center}
\end{table}

Table 1 shows p value scores for each category of the Godspeed questionnaire for formation one and two. The two tailed Mann-Whitney U test shows p values < 0.05 for the categories of Anthropomorphism, Animacy, and Perceived Intelligence, meaning they are statistically significant, and the null hypothesis can be rejected. The p values for these categories are well below 1 percent. These results can be explained as formation two seemed to follow the participants according to their movement. Regarding the categories of Likeability and Perceived Safety, they show p values > 0.05, so the null hypothesis cannot be rejected. However, these categories almost reached the 0.05 threshold needed to be statistically significant. The small drones were described as 'toy like' by several participants during the experiments. This description of the drones lead to similar Perceived Safety and likeability scores across these two formations.

\begin{figure}[htbp]
  \centering
  \includegraphics[width=\linewidth, height=2.5cm]{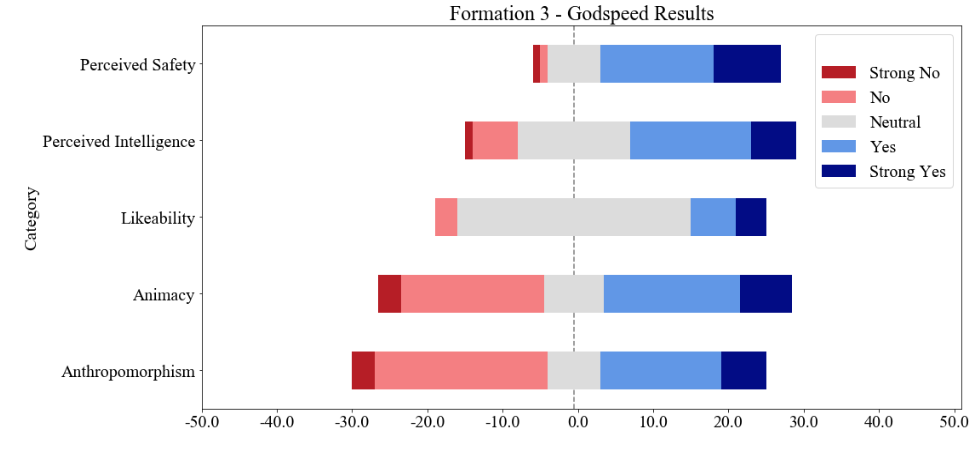}
  \caption{Tallied participant response for formation two - the arrow shape.}
\end{figure}

Figure 7 shows the participant responses to formation one, the arrow formation. The graph does not show the same extreme positive bias as Figure 6; however, it is arguably more positive than Figure 5 in the anthropomorphism category and less negative than Figure 5 in the animacy category. However, to clarify which if any differences are significant, we again apply the Mann-Whitney U test.
Table 2 shows p value scores for each category of the Godspeed questionnaire for formation one and three. The two tailed Mann-Whitney U test shows only Anthropomorphism received a p value < 0.05. All other categories received extraordinarily high p values. This result is significant as it suggests two possibilities, either the first formation is far too expressive and the seemingly randomly generated flight path is seen as intelligent/reactive or the third formation does not convey a clear message to be interpreted.

\begin{table}
    \caption{The results for the Mann-Whitney U comparing the formation one with formation three.}
    \begin{center}
    \begin{tabular}{|c|c|c|c|}
    \hline
        Category               & N  & P value  \\
        \hline
        Anthropomorphism       & 55 & 8.72E-03 \\
        Animacy                & 55 & 4.33E-01 \\
        Likeability            & 44 & 5.59E-01 \\
        Perceived Intelligence & 44 & 6.05E-01 \\
        Perceived Safety       & 33 & 2.89E-01 \\
        \hline
    \end{tabular}
    \end{center}
\end{table}

\subsection{Discussion}
Applying the Mann-Whitney U test shows for the categories of animacy, anthropomorphism, and perceived intelligence shows that the p-values less than .01, rejecting the null hypothesis that the formation is distinct in these categories. For the categories involving perceived safety and likeability the p-values greater than .05, meaning that the null hypothesis in these categories is not rejected. However, it must be noted that the p-values for these categories only narrowly missed the .05 values required to reject the null hypothesis, scoring .08 for likability and .11 for perceived safety. This result supports our thesis that a human participant is open to interpreting the motion of drones as having intent and potentially interpret their motion as communication. In this case, we designed the formation to convey the communication that a participant should not continue to move in their direction of motion. In the next section we will address the communication aspect. The arrow formation's means showed nearly no difference when compared to the random formation. The Mann-Whitney U test supports this claim as perceived safety, likeability, perceived intelligence, and animacy p-values greater than .05. Anthropomorphism's p-value less than .05 and was the only category that could have been distinguished when compared to the random formation.

Reaction time of the participants varied with each formation. The first formation had no meaning which led to the participants thinking about the formation for an average of 20 seconds. The second formation had a much more instantaneous reaction by the participants, many of them expressed that they understood what was going on one they started moving with the drones.

\subsection{Perceived Meaning}

Participant's drawings after each formation provide insight on what kinds of motions are considered intelligent and if there is a meaning communicated by the drone swarms. Even for the random swarm, most participants indicated that they saw some intelligence behind the swarm's movements according to the Perceived Intelligence category.

\begin{figure}[htbp]
  \centering
  \includegraphics[width=\linewidth, height=4cm]{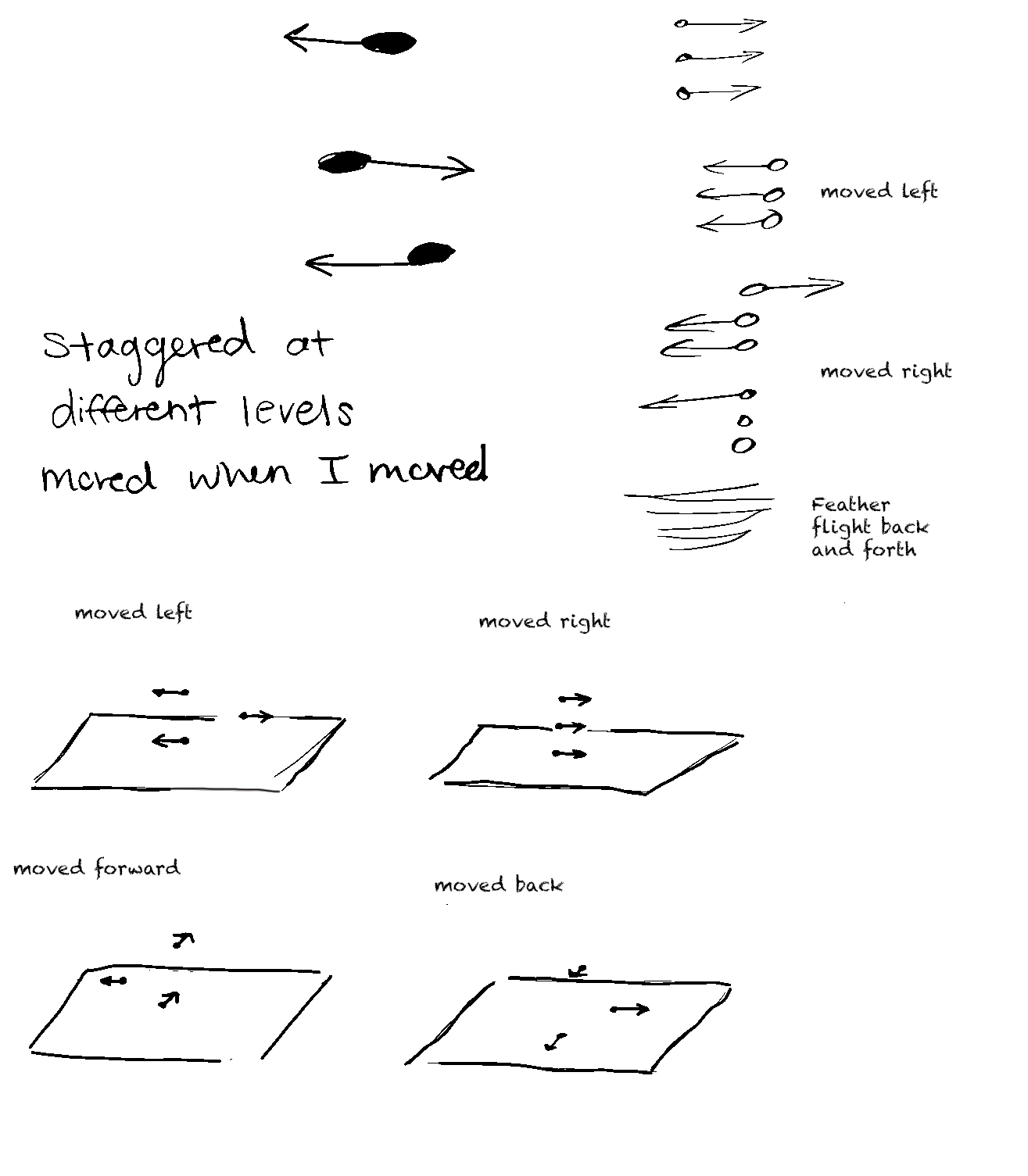}
  \caption{Retouched drawings done by participants of the random formation.}
\end{figure}

Participant drawings of the random swarm mainly depicted the swarm movement in phases (Fig. 8). A comment left on one of the drawings claiming that the drones "staggered ...when she moved" and the labeling of the when the participant stepped in a direction point to a belief that some participants believed the random movements of the drones related to their own movements. Short response answers asking to describe the swarm movement report that nearly all the participants felt that the drones followed them in some way. Only 3 participants reported that the movements were "calculated". When asked for the meaning behind the movements, 10 participants reported that there was no meaning behind the movements. The remaining participants reported that the drone's intent was to follow the participant with some suggesting "they were bees following a flower" and "they were the predators and I was the prey". The 'X' formation had the highest scored in the perceived intelligence category. All participants reported that this formation had some sort of meaning and intent.

\begin{figure}[htbp]
  \centering
  \includegraphics[width=\linewidth, height=3.2cm]{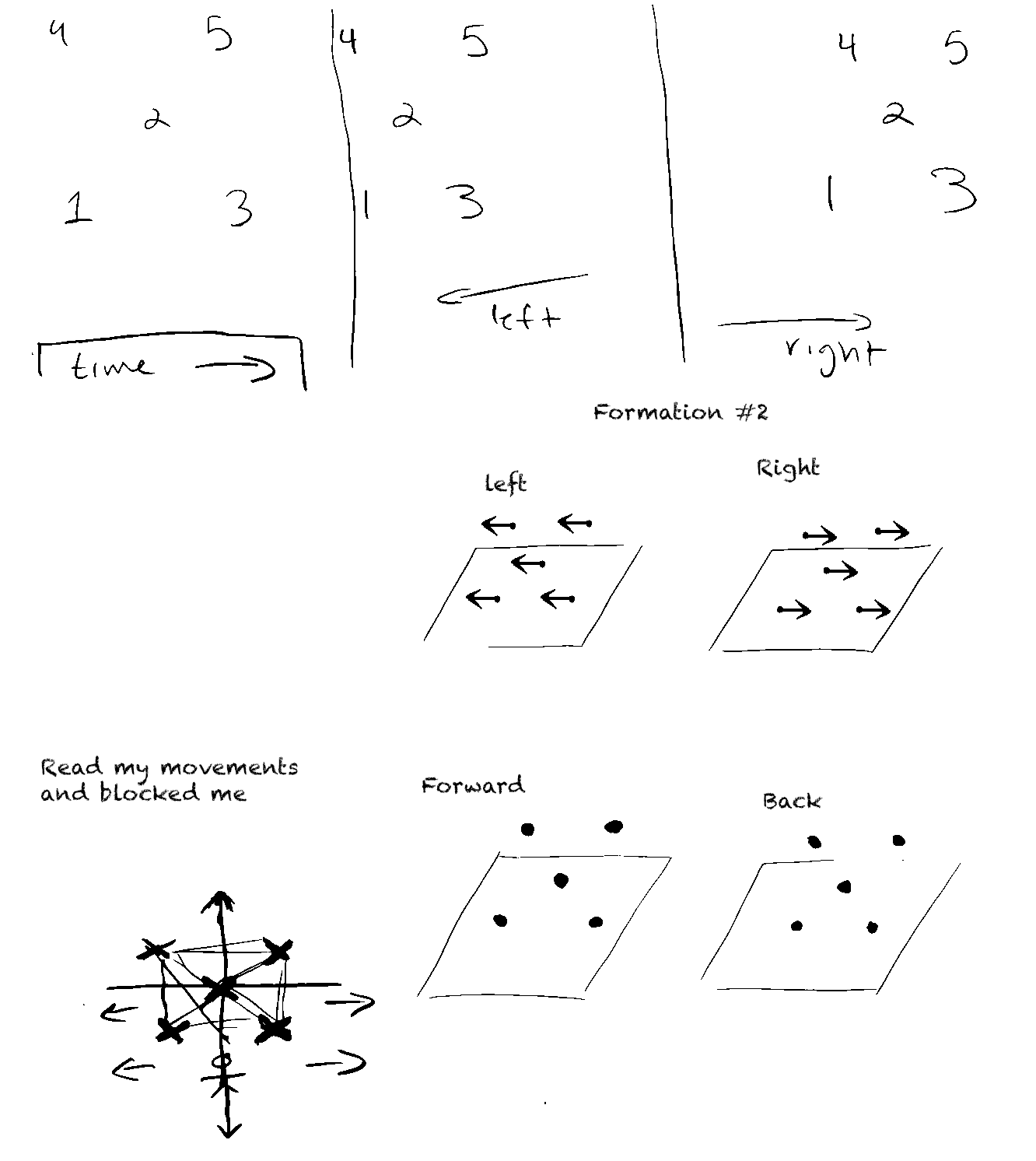}
  \caption{Retouched drawings done by participants of the 'X' formation}
\end{figure}

Just as for the random formation, the 'X' formation was drawn in phases by many participants. A comment left on one of the drawings claimed that the formation "read his movements and blocked" them and the corresponding movements of the participants and the drones suggest that the participants felt blocked by the formation. When asked to describe the motions, participants also commented the drones "rapidly" followed them based on their movement and "made an X shape". When asked about the drone formation's intent, 7 participants claimed the drones were trying to prevent them from moving forward claiming it was like "a wall" or "an X". 4 participants reported the drones were "like guards" or "soldiers", personifying the drones with a job related to authority. A participant reported that they felt "stalked" and compared the drones to an "evil cooperation following their movements". The amount of comparisons to real life entities correlates the high scores in categories animacy and anthropomorphism in the survey categories.

The arrow formation scores in perceived intelligence are very similar to the random formation and is not distinct from the random formation according to the Mann-Whitney U test. Only one third of the participants saw intent in this formation, and the remainder concluded the formation was generally a guide.

\begin{figure}[htbp]
  \centering
  \includegraphics[width=\linewidth, height=5.2cm]{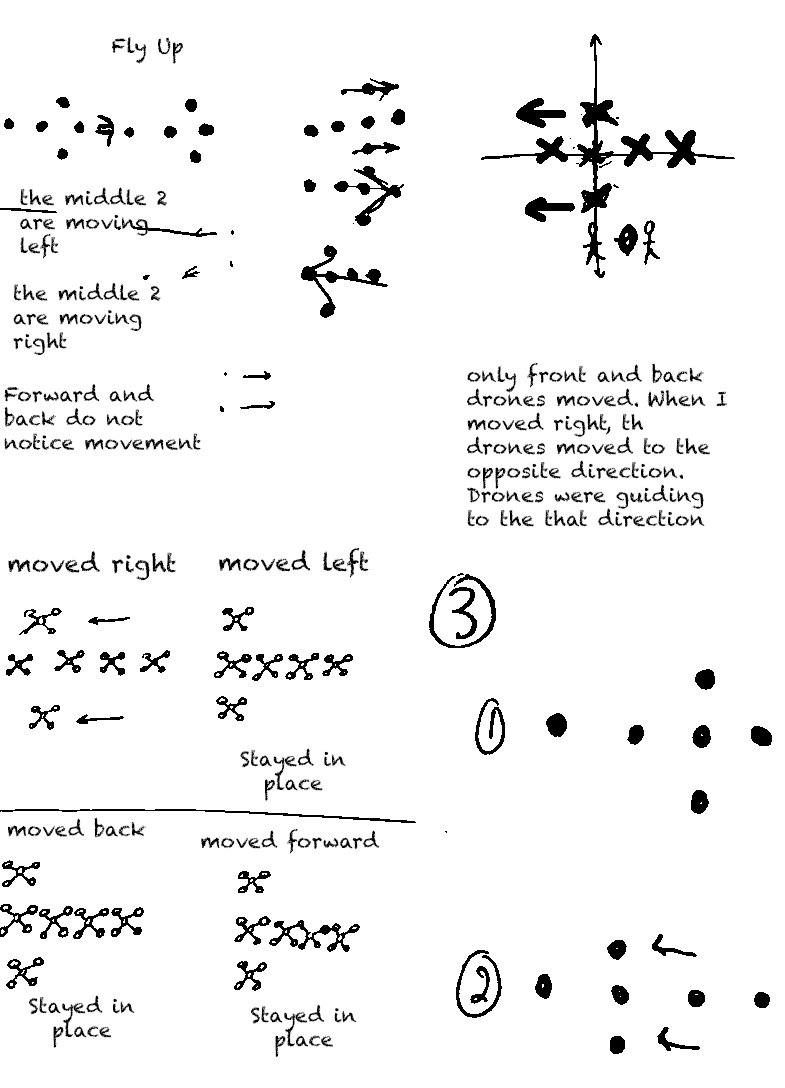}
  \caption{Retouched drawings done by participants of the arrow formation}
\end{figure}

Participant's comments describing the motion of this formation commented on the "limited" movement of the drones as this formation remained stationary for longer periods of time as appose to the random and the 'X' formation which moved more frequently (Figure 10). The participants reached 2 distinct meanings regarding the intent of the drone formation claiming the drones had no intent and the drones acted as a guide. Comments claiming the formation to be a guide compared the drones to a "crossing guard" and "a digital traffic sign". Participants who claimed there was no meaning, suggested the formation appeared to be a staircase or a random swarm of digital insets.

\section{Conclusion}

In this paper we report the first phase of our investigation of whether drone swarms can be used to communicate information to human observers on a range of different nonverbal channels. In particular, we report our experimental results for the questions of 1) whether a human observer would in fact consider a swarm of drones in their immediate vicinity to be non- threatening enough to be a vehicle for communication, 2) whether a human would in fact intuit some communication from the swarm behavior and 3) whether there was some predictability in what was intuited. We designed and implemented several swarm formations for swarms of Crazyflie drones -- small, light drones. We exposed participants to these formations, which included some swarm responses and audio responses to participant motions, and we administer an extended version of the Godspeed questionnaire.

Our results demonstrate that the drones were perceived as nonthreatening for all formations. Using the Mann-Whitney U test, we show that formation two, a formation in which the drones create an 'X' shape and beep when then participant approaches the 'X', is significantly different from formation one, random motion in perceived animacy and anthropomorphism. This supports our conclusion that a human participant is prepared to see the motion of the drones as having intent and thus is ready to communicate intent. To understand what was conveyed in the communication we also participants to draw and comment on what they felt was communicated. Comments for the 'X' formation included 'like a wall' or 'like guards' - qualities that we would argue support the notion of the swarm communicating the intent to block forward motion.
The experiment highlights the potential for a wide range of applications of drone swarms, leveraging the unique and key advantages of flexibility and mobility that drone user interfaces bring to the table.

The institution-mandated existence of a net in this work must be considered in interpreting our results. Our objective is to use an invisible barrier in future work. Our inclusion of a simple audio signal in addition to motion must also be considered: how much of a cue was taken from the sound as opposed to motion? Since natural swarms include motion and sound we argue the combination is important and consideration of motion alone departs from intuitive the basis behind our work. In future work we plan to evaluate a more application specific scenario of communication and further explore the potential for human-drone interaction.

\bibliographystyle{IEEEtran}
\bibliography{Ref1}

\begin{thebibliography}{10}
\providecommand{\url}[1]{#1}
\csname url@samestyle\endcsname
\providecommand{\newblock}{\relax}
\providecommand{\bibinfo}[2]{#2}
\providecommand{\BIBentrySTDinterwordspacing}{\spaceskip=0pt\relax}
\providecommand{\BIBentryALTinterwordstretchfactor}{4}
\providecommand{\BIBentryALTinterwordspacing}{\spaceskip=\fontdimen2\font plus
\BIBentryALTinterwordstretchfactor\fontdimen3\font minus
  \fontdimen4\font\relax}
\providecommand{\BIBforeignlanguage}[2]{{%
\expandafter\ifx\csname l@#1\endcsname\relax
\typeout{** WARNING: IEEEtran.bst: No hyphenation pattern has been}%
\typeout{** loaded for the language `#1'. Using the pattern for}%
\typeout{** the default language instead.}%
\else
\language=\csname l@#1\endcsname
\fi
#2}}
\providecommand{\BIBdecl}{\relax}
\BIBdecl

\bibitem{Armstrong1}
W.~C. Stokoe, D.~F. Armstrong, M.~A. Karchmer, and V.~C.~J. V., \emph{The study
  of signed languages: essays in honor of William C. Stokoe}.\hskip 1em plus
  0.5em minus 0.4em\relax Washington, D.C.: Gallaudet University Press, 2011.

\bibitem{craig2}
C.~J. Baxter, D.~Druckman, and R.~M. Rozelle, \emph{Nonverbal Communication:
  Survey, Theory, and Research}.\hskip 1em plus 0.5em minus 0.4em\relax NY:
  SAGE Publications. Inc, 1982.

\bibitem{darwin3}
C.~Darwin, \emph{The Expression of the Emotions in Man and Animals}.\hskip 1em
  plus 0.5em minus 0.4em\relax United Kingdom, London: John Murray, 1872.

\bibitem{nguyen4}
T.-H.~D. Nguyen, D.~Lyons, and K.~Grispino, ``Towards affective drone swarms in
  public spaces,'' in \emph{4th Workshop on Public Space HRI}.\hskip 1em plus
  0.5em minus 0.4em\relax Barcelona, Spain: PubRob, 2018.

\bibitem{bartneck5}
C.~Bartneck, D.~Kuli{\'{c}}, E.~Croft, and S.~Zoghbi, ``Measurement instruments
  for the anthropomorphism, animacy, likeability, perceived intelligence, and
  perceived safety of robots,'' \emph{Int. J. Soc. Robotics}, vol.~1, no.~1,
  pp. 71--81, Jan 2009.

\bibitem{nam6}
C.~Nam, P.~Walker, H.~Li, M.~Lewis, and K.~Sycara, ``Models of trust in human
  control of swarms with varied levels of autonomy,'' \emph{IEEE Trans.
  Human-Machine Sys.}, vol.~{}, no.~{}, February 2019.

\bibitem{8habib}
L.~{Habib}, M.~{Pacaux-Lemoine}, and P.~{Millot}, ``Human-robots team
  cooperation in crisis management mission,'' in \emph{IEEE Int. Conf. on
  SMC}.\hskip 1em plus 0.5em minus 0.4em\relax Miyazaki, Japan: IEEE, Oct 2018,
  pp. 3219--3224.

\bibitem{9Cauchard}
J.~R. Cauchard, J.~L. E, K.~Y. Zhai, and J.~A. Landay, ``Drone \&\#38; me: An
  exploration into natural human-drone interaction,'' in \emph{ACM Int. Joint
  Conf. on Pervasive and Ubiquitous Computing}, NY, USA, 2015, pp. 361--365.

\bibitem{10nagi}
J.~{Nagi}, A.~{Giusti}, G.~A.~D. {Caro}, and L.~M. {Gambardella}, ``Human
  control of uavs using face pose estimates and hand gestures,'' in \emph{2014
  9th ACM/IEEE Int. Conf. on HRI}, Bielefeld, Germany, March 2014, pp. 1--2.

\bibitem{13Burke}
J.~Burke, K.~S. Pratt, and R.~Murphy, ``Toward developing hri metrics for
  teams: Pilot testing in the field,'' in \emph{Proceedings of Metrics for
  Human-Robot Interaction}.\hskip 1em plus 0.5em minus 0.4em\relax Amsterdam,
  The Netherlands: ACM/IEEE Int. Conf. on HRI, 2008, pp. 21--28.

\bibitem{Carpinella17}
C.~Carpinella, A.~Wyman, M.~Perez, and S.~Stroessner, ``The robotic social
  attributes scale (rosas): Development and validation.''\hskip 1em plus 0.5em
  minus 0.4em\relax 2017 ACM/IEEE Int. Conf. on Human-Robot Interaction, 2017.

\bibitem{Nomura08}
T.~Nomura, T.~Kanda, T.~Suzuki, and K.~Kato, ``Prediction of human behavior in
  human--robot interaction using psychological scales for anxiety and negative
  attitudes,'' \emph{IEEE Trans. Rob.}, vol.~24, 2004.

\bibitem{14Saldien}
J.~Saldien, B.~Vanderborght, K.~Goris, M.~Damme, and D.~Lefeber, ``A motion
  system for social and animated robots,'' \emph{Int. J. Adv. Rob. Sys.},
  vol.~11, 05 2014.

\bibitem{15meng}
L.~Meng, D.~Lin, A.~Francey, R.~Gorbet, P.~Beesley, and D.~Kulić, ``Learning
  to engage with interactive systems: A field study,'' 2019.

\bibitem{new2}
M.~Santos and M.~Egerstedt, ``From motions to emotions: Can the fundamental
  emotions be expressed in a robot swarm?'' \emph{ArXiv}, vol. abs/1903.12118,
  2019.

\bibitem{new1}
F.~{Levillain}, D.~{St-Onge}, E.~{Zibetti}, and G.~{Beltrame}, ``More than the
  sum of its parts: Assessing the coherence and expressivity of a robotic
  swarm,'' in \emph{27th IEEE Int. Symp. on Robot and Human Interactive
  Communication (RO-MAN)}, Aug 2018, pp. 583--588.

\end{thebibliography}

\end{document}